\documentclass[11pt,a4paper]{article}
\usepackage[hyperref]{acl2021}
\usepackage{times}
\usepackage{latexsym}
\usepackage{url}
\usepackage{makecell}
\usepackage{multirow}
\usepackage{booktabs}
\usepackage{amsmath}
\usepackage{graphicx}
\usepackage{stfloats}
\usepackage{bm}
\usepackage[linesnumbered,ruled,lined]{algorithm2e}

\usepackage{paralist}

\makeatletter
\def\UrlAlphabet{%
      \do\a\do\b\do\c\do\d\do\e\do\f\do\g\do\h\do\i\do\j%
      \do\k\do\l\do\m\do\n\do\o\do\p\do\q\do\r\do\s\do\t%
      \do\u\do\v\do\w\do\x\do\y\do\z\do\A\do\B\do\C\do\D%
      \do\E\do\F\do\G\do\H\do\I\do\J\do\K\do\L\do\M\do\N%
      \do\O\do\P\do\Q\do\R\do\S\do\T\do\U\do\V\do\W\do\X%
      \do\Y\do\Z}
\def\UrlDigits{\do\1\do\2\do\3\do\4\do\5\do\6\do\7\do\8\do\9\do\0}
\g@addto@macro{\UrlBreaks}{\UrlOrds}
\g@addto@macro{\UrlBreaks}{\UrlAlphabet}
\g@addto@macro{\UrlBreaks}{\UrlDigits}
\makeatother

\aclfinalcopy
\def\aclpaperid{2428}

\usepackage{microtype}
\usepackage{amsthm}
\newtheorem{assumption}{Assumption}
\allowdisplaybreaks[4]

\everymath{\displaystyle}

\title{Provably Secure Generative Linguistic Steganography}

\author{Siyu Zhang, Zhongliang Yang, Jinshuai Yang, Yongfeng Huang \\
  Department of Electronic Engineering, Tsinghua University, Beijing 100084, China \\
  \texttt{\{zhangsiy19, yjs20\}@mails.tsinghua.edu.cn} \\
  \texttt{yangzl15@tsinghua.org.cn} \\
  \texttt{yfhuang@mail.tsinghua.edu.cn}}

\date{}

\begin{document}
\maketitle
\begin{abstract}
Generative linguistic steganography mainly utilized language models and applied steganographic sampling (stegosampling) to generate high-security steganographic text (stegotext). However, previous methods generally lead to statistical differences between the conditional probability distributions of stegotext and natural text, which brings about security risks. In this paper, to further ensure security, we present a novel provably secure generative linguistic steganographic method \textbf{ADG}, which recursively embeds secret information by \textbf{A}daptive \textbf{D}ynamic \textbf{G}rouping of tokens according to their probability given by an off-the-shelf language model. We not only prove the security of ADG mathematically, but also conduct extensive experiments on three public corpora to further verify its imperceptibility. The experimental results reveal that the proposed method is able to generate stegotext with nearly perfect security.
\end{abstract}

\section{Introduction}

Steganography is the technology of hiding secret information within an innocent natural carrier (such as image \cite{hussain2018image}, audio \cite{mishra2018audio}, video \cite{liu2019video}, text \cite{krishnan2017overview}, etc) in order to avoid eavesdropping. Steganography differs from cryptography in that cryptography only conceals the content of secret information, whereas steganography even conceals its very existence, which makes it more secure and reliable in some scenarios \cite{anderson1998limits}. 

Natural language is suitable as a carrier of steganography by virtue of its high robustness in transmission \cite{ziegler2019neural}. Unlike digital images or digital audio which is sensitive to distortions like compression, cropping, blurring or pixel-wise dropout, text can usually be transmitted losslessly through different kinds of public channels. Nevertheless, text generally has low entropy and lacks sufficient redundancy for information hiding \cite{sharma2016analysis}, which often results in low embedding capacity of linguistic steganography. For example, in traditional modification-based methods (such as synonym substitution \cite{xiang2014linguistic, xiang2018reversible} and spelling transformation \cite{shirali2008text}), where secret information is encoded by slightly modifying an existing covertext, the options for modification can be very limited to keep the text fluent enough so as not to arouse suspicions.

In recent years, powered by the advanced technology of deep learning and natural language processing, language models based on neural networks have made significant progress in generating fluent text \cite{radford2019language, brown2020language}, which bring new vitality to linguistic steganography and facilitate the investigation of generation-based methods \cite{fang2017generating, yang2018rnn, dai2019towards, ziegler2019neural, yang2020vae, zhou2021linguistic}. The generative linguistic steganography directly transform secret information into innocuous-looking steganographic text (stegotext) without any covertext. Using an off-the-shelf language model, secret information can be encoded in the selection of token at each time step autoregressively during the generation procedure, which greatly alleviates the drawback of low embedding capacity. However, previous methods inevitably introduce distortions during generation. The imperceptibility of generative linguistic steganography still needs further optimization.

In this paper, we aim to further improve the imperceptibility of generative linguistic steganography. The contributions of this work are the following:
\begin{enumerate}
\item We present \textbf{ADG} (\textbf{A}daptive \textbf{D}ynamic \textbf{G}rouping), a novel generative linguistic steganographic method based on off-the-shelf language models, which groups the tokens adaptively in accordance with their probability at each time step to embed secret information dynamically in the generated stegotext.
\item We discuss the security of ADG and give mathematical proof, which reveals that the proposed method is provably secure.
\item Through quantitative analysis, we derive satisfactory experimental results in terms of both imperceptibility and embedding capacity, which further verifies the effectiveness of ADG.
\end{enumerate}
Our code is available at \url{https://github.com/Mhzzzzz/ADG-steganography}.

\section{Formalism}
\subsection{Notation}
We use lowercase letters in bold type (e.g. $\bm{a}$) to denote vectors, normal lowercase letters (e.g. $a$) to denote scalars and uppercase letters (e.g. $A$) to denote sets. We use the symbol $|A|$ to denote the size of a set. Calligraphic letters denote neural models (e.g. $\mathcal{A}$). Both English letters and Greek letters are adopted. We use $p(\cdot)$ and $q(\cdot)$ to denote distributions and $f(\cdot)$ to denote functions, which are usually shortened to $p$, $q$ and $f$. Subscripts and superscripts are used to tell the different variables/distributions/functions apart.

\subsection{Generative Linguistic Steganography}
Language modeling is a task to estimate the joint distribution of serialized natural language $p_{LM}(\bm{w})$, where $\bm{w}$ is a sequence of $n$ tokens $[w_1, w_2, ..., w_n]$ and each token belongs to the vocabulary $\Sigma$. For an autoregressive language model $\mathcal{L}$, the output is usually factorized as a product of conditional distribution of the current token 
\begin{equation}
  \begin{aligned}
    p_{LM}(\bm{w}) &= p_{LM}(w_1, w_2, ..., w_{n}) \\
    &= p_{LM}(w_1) \cdot \prod \limits_{t=2}^{n}{p_{LM}(w_t|w_1,...,w_{t-1})}.
  \end{aligned}
\end{equation}

According to \citet{simmons1984prisoners}, it is usually supposed that Alice (sender) wants to send a secret message $\bm{m} \sim  \mbox{Uniform}(\{0,1\}^l)$ to Bob (receiver) through a public channel monitored by Eve (adversary). In generative linguistic steganography, they share an {\it embedding} algorithm $f_{emb}$ which takes a language model $\mathcal{L}$ and the secret message $\bm{m}$ as input and then outputs {\it stegotext} $\bm{y}$ to transmit. They also share a corresponding {\it extraction} algorithm $f_{ext}$, which is the inverse mapping of $f_{emb}$ that is able to recover the secret message $\bm{m}$ according to the language model $\mathcal{L}$ and the received stegotext $\bm{y}$.

\subsection{Imperceptibility}
\label{subsection imperceptibility}
In order to avoid raising Eve's suspicions, stegotext $\bm{y}$ is required to be fluent enough and statistically indistinguishable from natural innocuous text $\bm{x}$, which we call {\it covertext}. \citet{cachin1998information} proposed the {\it information-theoretic security} of steganography to measure the statistical imperceptibility quantitatively, which is defined as the Kullback-Leibler divergence (KL divergence) between the distributions of covertext $\bm{x}$ and stegotext $\bm{y}$. The distortion of generative linguistic steganography is two-fold: one is introduced by the bias of the language models, which is the gap between the true distribution of natural text $p_{true}(\bm{x})$ and the modeled distribution $p_{LM}(\bm{x})$; the other is introduced by $f_{emb}$. Instead of directly sampling from the modeled distribution, the embedding algorithm $f_{emb}$ actually provides a special way to sample from $p_{LM}(\bm{y})$, which we call steganographic sampling (stegosampling). It is equivalent to sampling from a modified distribution $q(\bm{y})$ produced by an implicit language model $\mathcal{L}'$. In a word, the latter distortion is the gap between $p_{LM}(\bm{y})$ and $q(\bm{y})$, which can also be regarded as the gap between the conditional distributions $p_{LM}(y_t|y_{<t})$ and $q(y_t|y_{<t})$. We simply use $p_{LM}$ and $q$ to refer to the conditional distributions in the rest of this paper.

\section{Related Work}
In the early stage, some researchers investigated rule-based approaches or using Markov Chains to achieve generative linguistic steganography \cite{wayner1992mimic, chapman1997hiding, chapman2001practical, chapman2002plausible, dai2010text, moraldo2014approach, luo2016text, yang2018automatically}. However, these methods followed a simplistic pattern and are hard to guarantee the grammatical correctness and the semantic fluency of the generated stegotext.

With the development of deep learning, language models based on neural networks show great performance on automatic text generation. The pattern of generating stegotext with neural language models has been widely accepted.
\citet{fang2017generating} proposed a linguistic steganographic method that randomly partitioned the vocabulary $\Sigma$ into $2^{b}$ bins $[B_1, B_2, ..., B_{2^{b}}]$ and each one contained $|\Sigma| / 2^{b}$ tokens. At each time step, they selected the token with the highest probability within the bin according to the $b-$bit secret information to be embedded.
\citet{yang2018rnn} improved the embedding algorithm by building the mapping from secret information to tokens dynamically at each time step rather than statically in advance. Concretely, the top $2^k$ tokens with the highest probability were encoded by Huffman coding algorithm. Then they took the token which has the same code as the secret information.
\citet{dai2019towards} proposed {\it patient-Huffman}, which was an improved version of \citet{yang2018rnn} that sacrificed embedding capacity for imperceptibility. They first calculated the distortion (total variation distance or KL divergence) between $q$ and $p_{LM}$ and then only used Huffman coding embedding algorithm to embed secret information when the distortion was less than a preset threshold $\delta$. Otherwise they directly sampled a token to avoid high distortion occasions.
\citet{ziegler2019neural} employed arithmetic coding to embed secret information. They truncated the top $h$ likely tokens and left out the low-probability long-tails. Then the tokens are encoded by arithmetic coding algorithm and selected according to the secret information. Compared with other coding algorithm, arithmetic coding has higher compression rate, which results in less damage to conditional probability distribution $p_{LM}$ and helps to improve imperceptibility.



\section{ADG Methodology}
According to the analysis in Section \ref{subsection imperceptibility}, the distortion of generative linguistic steganography includes the bias of the language model $\mathcal{L}$ and the damage to the conditional distribution caused by the embedding algorithm $f_{emb}$. The former is not our research priority. With the development of automatic text generation, the former distortion can be gradually minimized. In this paper, we mainly pay attention to the latter distortion. We aim to seek an optimal solution theoretically and experimentally.

Given an off-the-shelf language model, how can we embed secret information to the generated tokens? Unlike previous works that encoded the conditional distribution by lossless coding algorithm, we achieve this goal in a novel way by grouping. Through mathematical analysis and proof, we propose a provably secure method ADG, which does little damage to the conditional distribution and is nearly equivalent to directly sampling from the full distribution. In this section, we investigate the security of steganography by grouping and give detailed descriptions of the proposed method.

\subsection{Steganography by Grouping}
Steganography by grouping is to group all tokens in the vocabulary into several groups, so that each group represents a unique secret message. E.g. we can  Tokens belonging to the target group are able to make up the stegotext. In such a way, Bob reads each token in the sequence in turn and performs the same grouping operation to extrapolate which groups the current token belongs to, thereby extracting the corresponding secret information. The key question is: how to group the tokens at each time step to ensure an optimal imperceptibility? We have the following assumption.

\begin{assumption}
For secret information in the form of uniformly distributed bitstream, adaptively grouping the vocabulary into $u$ groups ($u = 2^r, r \in N, r \leq \log_2{|\Sigma|}$) with equal probability will ensure the optimal imperceptibility.
\end{assumption}

\begin{proof}
Assuming that the discrete conditional probability distribution $p_{LM}$ is arbitrarily partitioned into $u$ groups to embed $r$-bit secret information. $p_{ij}$ denotes the probability of the $j$-th token in the $i$-th group. $\eta_i$ and $n_i$ denote the total probability and the size of the $i$-th group respectively. Then we have
\begin{equation}
\label{equation1}
	\sum_{j=1}^{n_i}{p_{ij}=\eta_i}, \	\sum_{i=1}^{u}{\eta_i=1}.
\end{equation}

Our goal is to figure out the grouping algorithm to achieve the best imperceptibility, i.e. to minimize the gap between $p_{LM}$ and $q$. First of all, starting from the modeled distribution $p_{LM}=[..., p_{ij}, ...]$, we calculate the equivalent distribution $q$. The probability of each token is firstly normalized within its group ($1/\eta_i$) and then multiplied by the selected probability of the group, which is $1/u$ since secret information is uniformly distributed. Therefore, $q$ has the following form
\begin{equation}
	q=[..., p_{ij}/u\eta_i, ...].
\end{equation}
We measured the gap between the two distributions with KL divergence, which is
\begin{align}
	D_{KL}(p_{LM} || q) & = \sum{p_{LM} \log \frac{p_{LM}}{q}} \notag \\
	& = \sum_{i=1}^{u}{\sum_{j=1}^{n_i}{p_{ij} \log \frac{p_{ij}}{p_{ij}/u\eta_i}}} \notag \\
	& = \sum_{i=1}^{u}{\sum_{j=1}^{n_i}{p_{ij} \log (u\eta_i)}} \notag \\
	& = \sum_{i=1}^{u}{\log (u\eta_i) \sum_{j=1}^{n_i}{p_{ij}}} \notag \\
	& = \sum_{i=1}^{u}{\eta_i \log (u\eta_i)}.
\end{align}
Therefore, the KL divergence between the two distributions is a function of the vector $\bm{\eta} = [\eta_1, \eta_2, ..., \eta_u]$. 

Next, we will prove Assumption 1 in two steps.

\textbf{[1].} Considering the auxiliary function $f_{aux}(\eta) = \eta \log (u\eta), \ (0 \le \eta \le 1)$, we firstly analyse its concavity and convexity on the domain of definition. For every $\eta_1, \eta_2 \in (0, 1)$ and $0\leq \lambda \leq 1$,
\begin{equation}
  \begin{aligned}
    & f_{aux}(\lambda \eta_1 + (1 - \lambda) \eta_2) \\
    & -\lambda f_{aux}(\eta_1) - (1 - \lambda) f_{aux}(\eta_2) \\
    & = (\lambda \eta_1 + (1 - \lambda) \eta_2) \log (u(\lambda \eta_1 + (1 - \lambda) \eta_2)) \\
    & - \lambda(\eta_1 \log (u\eta_1)) - (1- \lambda)(\eta_2 \log (u\eta_2)) \\
    & = \lambda \eta_1 \log \frac{\lambda \eta_1 + (1 - \lambda) \eta_2}{\eta_1} \\
    & + (1 - \lambda)\eta_2 \ log \frac{\lambda \eta_1 + (1 - \lambda) \eta_2}{\eta_2} \\
    & \leq \lambda (\eta_1 (\frac{\lambda \eta_1 + (1 - \lambda) \eta_2}{\eta_1} - 1)) \\
    & + (1 - \lambda)\eta_2 (\frac{\lambda \eta_1 + (1 - \lambda) \eta_2}{\eta_2} - 1) \\
    & = \lambda (\lambda \eta_1 + (1 - \lambda) \eta_2 - \eta_1) \\
    & + (1 - \lambda)(\lambda \eta_1 + (1 - \lambda) \eta_2 - \eta_2) \\
    & = \lambda(\lambda - 1)\eta_1 + \lambda(1 - \lambda)\eta_2 \\
    & + (1 - \lambda)\lambda\eta_1 - (1 - \lambda)\lambda \eta_2 \\
    & = 0. \\
  \end{aligned}
\end{equation}
As a result, $f_{aux}(\eta)$ is convex over $(0, 1)$.

\textbf{[2].}  Then, when generalizing to $u$ variables $\eta_1, \eta_2, ..., \eta_u, \ \sum_{i=1}^{u}{\eta_i} = 1$, according to Jensen's inequality \cite{jensen1906fonctions}, there is
\begin{equation}
  \begin{aligned}
  & \frac{\sum_{i=1}^{u}{f_{aux}(\eta_i)}}{u} = \frac{\sum_{i=1}^{u}{\eta_i \log (u\eta_i)}}{u} \\
  & \geq f_{aux}(\frac{\sum_{i=1}^{u}{\eta_i}}{u}) = \frac{1}{u} \sum_{i=1}^{u}{\eta_i} \log (\sum_{i=1}^{u}{\eta_i}) = 0.
  \end{aligned}
\end{equation}
The equality sign holds if and only if 
\begin{equation}
\eta_1 = \eta_2 = ... = \eta_u.
\end{equation}
It means that $D_{KL}(p_{LM} || q) = \sum_{i=1}^{u}{f_{aux}(\eta_i)}$ takes the minimum value 0 when each component of $\bm{\eta}$ is equal, in which case $p_{LM}$ and $q$ are equivalent and that achieves the optimal information-theoretic security defined by \citet{cachin1998information}.
\end{proof}

Therefore, we basically construct the idea of our embedding algorithm,  that is, to adaptively group the vocabulary into multiple groups at each time step, so that each group is assigned approximately the same probability. In practice, since the probability distribution is discrete, the probability of groups may not be absolutely equal. Firstly, we determine the number of groups $u$ to be  its maximum value $2^{\lfloor - \log_2 {p_{max}} \rfloor}$, where $p_{max}$ is the highest probability in $p_{LM}$. Secondly, since the time complexity of solving the global optimal solution of equal grouping is unacceptable, we implement a suboptimal solution in ADG, as demonstrated in Algorithm \ref{algorithm even}. In line 10, we employ binary search algorithm to select the token that has the nearest probability of a given value. Our implementation enables us to obtain a unique grouping result for any $p_{LM}$, which ensures that the secret information can be extracted accurately and completely at the receiving end.

\begin{algorithm}[h]
  \caption{Suboptimal solution of equal grouping.}
  \label{algorithm even}
  \SetAlgoLined
    \KwData{vocabulary $\Sigma$, distribution $p_{LM}$}
    \KwResult{set of groups $G$}
    $\mbox{list of tokens} = $ sorted $(p_{LM})$\;
    $p_{max} = \mbox{probability of the first token}$\;
    $u = 2^{\lfloor - \log_2 {p_{max}}\rfloor}$\;
    $\mbox{mean} = 1/u$\;
    \For {($i=1; i \leq u-1; i$ ++)}{
    		$G_i = [\mbox{the first token}]$\;
    		remove the first token\;
	    \While{$\sum{\mbox{probability of }G_i} < \mbox{mean}$}{
	    		$\varepsilon = \mbox{mean} - \sum{\mbox{probability of }G_i}$\;
	    		select a token with the nearest probability of $\varepsilon$\;
	    		\If{$\mbox{probability of the token} - \varepsilon < \varepsilon$}{
	    			append the token to $G_i$\;
	    			remove the token\;
	    		}
	    		\Else{
	    			break\;
	    		}
	 	}
	 	$\mbox{mean} = \frac{\mbox{probability of the rest tokens}}{u - i}$;
	}
	append the rest tokens to $G_u$\;
	$G = [G_1, G_2, ..., G_u]$\;
\end{algorithm}

\begin{table*}[h]	
  \small
  \centering
  \caption{Results of ER, KLD$_1$ and KLD$_2$.}	
  \label{table result kl}
  \begin{tabular}{m{100pt}<{\raggedright} | m{22pt}<{\centering} m{27pt}<{\centering} m{27pt}<{\centering} | m{22pt}<{\centering} m{27pt}<{\centering} m{27pt}<{\centering} | m{22pt}<{\centering} m{27pt}<{\centering} m{27pt}<{\centering} }
  \toprule[1.2pt]
  \multicolumn{1}{c|}{\multirow{2}{*}{\textbf{METHOD}}} & \multicolumn{3}{c|}{\textbf{Movie}} & \multicolumn{3}{c|}{\textbf{News}} & \multicolumn{3}{c}{\textbf{Tweet}} \\
  \cline{2-10} \specialrule{0em}{2pt}{1pt}
  	  &	\textbf{$\uparrow$ER} & \textbf{$\downarrow$KLD$_1$} & \textbf{$\downarrow$KLD$_2$} &	\textbf{$\uparrow$ER} & \textbf{$\downarrow$KLD$_1$} & \textbf{$\downarrow$KLD$_2$} &	\textbf{$\uparrow$ER} & \textbf{$\downarrow$KLD$_1$} & \textbf{$\downarrow$KLD$_2$}  \\
  \midrule[0.9pt]
  Bins ($b=1$) &1.000 & 2.497 & 27.595 & 1.000 & 2.742 & 26.331 & 1.000 & 2.431 & 19.519 \\
  Bins ($b=2$) &2.000 & 2.338 & 33.206 & 2.000 & 2.593 & 35.207 & 2.000 & 2.421 & 17.604 \\
  Bins ($b=3$) &3.000 & 2.319 & 29.778 & 3.000 & 2.592 & 55.781 & 3.000 & 2.429 & 21.286 \\
  Bins ($b=4$) &4.000 & 2.439 & 54.155 & 4.000 & 2.550 & 87.441 & 4.000 & 2.314 & 27.230 \\
  Bins ($b=5$) &5.000 & 2.503 & 73.075 & 5.000 & 2.500 & 116.857 & 5.000 & 2.482 & 29.171 \\
  \hline \specialrule{0em}{1pt}{1pt}
  Huffman ($k=1$) &1.000 & 1.961 & 21.219 & 1.000 & 2.338 & 11.226 & 1.000 & 2.121 & 6.252 \\
  Huffman ($k=2$) &1.824 & 1.433 & 13.199 & 1.824 & 1.751 & 8.793 & 1.841 & 1.586 & 5.208 \\
  Huffman ($k=3$) &2.509 & 1.106 & 8.487 & 2.518 & 1.372 & 6.855 & 2.595 & 1.145 & 4.141 \\
  Huffman ($k=4$) &3.145 & 0.819 & 6.334 & 3.224 & 1.084 & 5.419 & 3.266 & 0.880 & 3.197 \\
  Huffman ($k=5$) &3.705 & 0.658 & 4.657 & 3.872 & 0.838 & 3.995 & 3.932 & 0.694 & 2.738 \\
  \hline \specialrule{0em}{1pt}{1pt}
  Patient-Huffman ($\delta=1.0$) &1.125 & 0.327 & \textbf{0.767} & 0.809 & 0.256 & \textbf{0.441} & 0.988 & 0.298 & \textbf{0.545} \\
  Patient-Huffman ($\delta=1.5$) &1.711 & 0.588 & 2.132 & 1.460 & 0.559 & 1.817 & 1.668 & 0.621 & 1.280 \\
  Patient-Huffman ($\delta=2.0$) &2.129 & 0.819 & 4.564 & 1.905 & 0.808 & 3.497 & 2.201 & 0.908 & 2.445 \\
  \hline \specialrule{0em}{1pt}{1pt}
  Arithmetic ($h=100$) &4.224 & 0.362 & 2.956 & 4.412 & 0.425 & 2.269 & 4.308 & 0.333 & 1.508 \\
  Arithmetic ($h=200$) &4.651 & 0.240 & 2.321 & 4.908 & 0.295 & 1.688 & 4.805 & 0.253 & 1.749 \\
  Arithmetic ($h=300$) &4.903 & 0.205 & 1.903 & 5.127 & 0.245 & 1.426 & 4.942 & 0.206 & 1.242 \\
  \hline
  \hline \specialrule{0em}{1pt}{1pt}
  ADG &\textbf{5.147} & \textbf{0.033} & 1.946 & \textbf{5.650} & \textbf{0.027} & 0.866 & \textbf{5.411} & \textbf{0.048} & 1.189 \\
  \bottomrule[1.2pt]
  \end{tabular}
\end{table*}

\subsection{Recursion and Pruning}
After obtaining the grouping results, we can select the group according to the next $\log u$ bits of secret information to be embedded and simply sample a token in the group to generate stegotext. As a matter of fact, we can also continue grouping the obtained groups to further enlarge the embedding capacity and recursively grouping the new groups until it is impossible to be equally participated (the normalized $p_{max}$ of the current group is greater than 0.5). In order to improve the efficiency of the recursive grouping, we employ pruning strategy to remove the redundant grouping operations. We only need to recursively group the selected groups every time in accordance with the secret information to be embedded. In this manner, the amount of secret information embedded in each token is adjusted dynamically according to its probability distribution.

To sum up, at each time step, the proposed ADG embedding algorithm first conducts the equal grouping algorithm adaptively according to the conditional distribution, and then recursively repeats the operation on the selected group dynamically according to the secret information, until it is indivisible. At last, we normalize the probability of the last selected group and sample a token to generate the stegotext. We have proved the security of equal grouping algorithm. Obviously, it can also be extended to the recursive manner of ADG, which means the proposed method is provably secure. 

\subsection{Information Extraction}
The extraction algorithm is basically the inverse process of the embedding algorithm. For an exactly successful extraction, Alice and Bob have to share the same language model, vocabulary and grouping algorithm. At each time step, Bob is supposed to recursively operate the same grouping algorithm as Alice do, and then select the group contains the current token in the stegotext. The index of the selected groups reveal the embedded secret information.

\begin{table*}[h]	
  \small
  \centering
  \caption{Results of ER, Acc$_1$ and Acc$_2$.}	
  \label{table result acc}
  \begin{tabular}{m{100pt}<{\raggedright} | m{22pt}<{\centering} m{27pt}<{\centering} m{27pt}<{\centering} | m{22pt}<{\centering} m{27pt}<{\centering} m{27pt}<{\centering} | m{22pt}<{\centering} m{27pt}<{\centering} m{27pt}<{\centering} }
  \toprule[1.2pt]
  \multicolumn{1}{c|}{\multirow{2}{*}{\textbf{METHOD}}} & \multicolumn{3}{c|}{\textbf{Movie}} & \multicolumn{3}{c|}{\textbf{News}} & \multicolumn{3}{c}{\textbf{Tweet}} \\
  \cline{2-10} \specialrule{0em}{2pt}{1pt}
  	  &	\textbf{$\uparrow$ER} & \textbf{Acc$_1$} & \textbf{Acc$_2$} &	\textbf{$\uparrow$ER} & \textbf{Acc$_1$} & \textbf{Acc$_2$} &	\textbf{$\uparrow$ER} & \textbf{Acc$_1$} & \textbf{Acc$_2$}  \\
  \midrule[0.9pt]
  Bins ($b=1$) &1.000 & 0.873 & 0.854 & 1.000 & 0.887 & 0.856 & 1.000 & 0.787 & 0.814 \\
  Bins ($b=2$) &2.000 & 0.812 & 0.802 & 2.000 & 0.855 & 0.830 & 2.000 & 0.739 & 0.753 \\
  Bins ($b=3$) &3.000 & 0.810 & 0.789 & 3.000 & 0.833 & 0.819 & 3.000 & 0.720 & 0.733 \\
  Bins ($b=4$) &4.000 & 0.825 & 0.832 & 4.000 & 0.843 & 0.852 & 4.000 & 0.748 & 0.760 \\
  Bins ($b=5$) &5.000 & 0.876 & 0.872 & 5.000 & 0.877 & 0.882 & 5.000 & 0.750 & 0.786 \\
  \hline \specialrule{0em}{1pt}{1pt}
  Huffman ($k=1$) &1.000 & 0.891 & 0.891 & 1.000 & 0.891 & 0.885 & 1.000 & 0.785 & 0.806 \\
  Huffman ($k=2$) &1.824 & 0.838 & 0.836 & 1.824 & 0.851 & 0.826 & 1.841 & 0.749 & 0.758 \\
  Huffman ($k=3$) &2.509 & 0.796 & 0.760 & 2.518 & 0.816 & 0.785 & 2.595 & 0.684 & 0.702 \\
  Huffman ($k=4$) &3.145 & 0.713 & 0.690 & 3.224 & 0.768 & 0.718 & 3.266 & 0.634 & 0.632 \\
  Huffman ($k=5$) &3.705 & 0.673 & 0.645 & 3.872 & 0.710 & 0.664 & 3.932 & 0.602 & 0.593 \\
  \hline \specialrule{0em}{1pt}{1pt}
  Patient-Huffman ($\delta=1.0$) &1.125 & 0.588 & 0.578 & 0.809 & 0.559 & 0.542 & 0.988 & 0.528 & 0.552 \\
  Patient-Huffman ($\delta=1.5$) &1.711 & 0.654 & 0.683 & 1.460 & 0.674 & 0.683 & 1.668 & 0.589 & 0.581 \\
  Patient-Huffman ($\delta=2.0$) &2.129 & 0.722 & 0.714 & 1.905 & 0.725 & 0.726 & 2.201 & 0.650 & 0.661 \\
  \hline \specialrule{0em}{1pt}{1pt}
  Arithmetic ($h=100$) &4.224 & 0.601 & 0.582 & 4.412 & 0.630 & 0.608 & 4.308 & 0.547 & 0.554 \\
  Arithmetic ($h=200$) &4.651 & 0.565 & 0.556 & 4.908 & 0.594 & 0.559 & 4.805 & 0.537 & 0.561 \\
  Arithmetic ($h=300$) &4.903 & 0.571 & 0.562 & 5.127 & 0.558 & 0.566 & 4.942 & 0.532 & 0.534 \\
  \hline
  \hline \specialrule{0em}{1pt}{1pt}
  ADG &\textbf{5.147} & \textbf{0.548} & \textbf{0.544} & \textbf{5.650} & \textbf{0.543} & \textbf{0.519} & \textbf{5.411} & \textbf{0.496} & \textbf{0.497} \\
  \bottomrule[1.2pt]
  \end{tabular}
\end{table*}

\section{Experimental Results and Analysis}
\label{section experimental results and analysis}
In this section, we evaluate the performance of ADG in terms of both embedding capacity and imperceptibility. Details of our experiments and the analysis of the results are present in the following subsections.

\subsection{Datasets}
We evaluated the performance of ADG on three public corpora, namely ``Large Movie Review Dataset'' (\textbf{Movie}) \cite{maas2011learning}, ``All the News'' (\textbf{News})\footnote{\url{https://www.kaggle.com/snapcrack/all-the-news}} and ``Sentiment140'' (\textbf{Tweet}) \cite{go2009twitter}. Large movie review dataset is originally built for binary sentiment classification, containing 100,000 movie reviews in total crawled from IMDb\footnote{\url{https://www.imdb.com/}}. ``All the news'' is a collection of publications of mainstream news media. Sentiment140 is also used in sentiment analysis tasks, which contains 1,600,000 tweets extracted from Twitter\footnote{\url{https://twitter.com/}}.

\begin{table}[h]	
  \small
  \centering
  \caption{Statistics of the preprocessed datasets.}	
  \label{table dataset}
  \begin{tabular}{m{40pt}<{\centering} m{30pt}<{\centering} m{50pt}<{\centering} m{50pt}<{\centering}}
  	\toprule[1.2pt]
  	\textbf{DATASET} & \textbf{$|\bm{\Sigma}|$} & \textbf{$|$TRAINING$|$} & \textbf{$|$TEST$|$} \\
  	\midrule[0.9pt]
  	Movie & 37,800 & 1,002,609 & 111,402 \\
  	News & 50,178 & 1,461,567 & 162,397 \\
  	Tweet & 30,152 & 1,572,599 & 174,734 \\
  	\bottomrule[1.2pt]
  \end{tabular}
\end{table}

We converted the raw text to lowercase and removed HTML tags and most punctuations, then segmented it into sentences with NLTK tools \cite{loper2002nltk}. We filtered out sentences with length below 5 or above 200. For the convenience of training and evaluation, any token occurring less than 10 times was mapped to a special token ``\_UNK''. We also added ``\_BOS'' and ``\_EOS'' at the beginning and end of each sentence to help training. Sentences in a batch were padded to the same length with a special padding token ``\_PAD''. Finally, we divided the preprocessed corpora into training set and test set according to the ratio of 9:1. Statistics are demonstrated in Table \ref{table dataset}.

\begin{table*}[h]	
  \small
  \centering
  \caption{Results of ER, EER$_1$ and EER$_2$.}	
  \label{table result eer}
  \begin{tabular}{m{100pt}<{\raggedright} | m{22pt}<{\centering} m{27pt}<{\centering} m{27pt}<{\centering} | m{22pt}<{\centering} m{27pt}<{\centering} m{27pt}<{\centering} | m{22pt}<{\centering} m{27pt}<{\centering} m{27pt}<{\centering} }
  \toprule[1.2pt]
  \multicolumn{1}{c|}{\multirow{2}{*}{\textbf{METHOD}}} & \multicolumn{3}{c|}{\textbf{Movie}} & \multicolumn{3}{c|}{\textbf{News}} & \multicolumn{3}{c}{\textbf{Tweet}} \\
  \cline{2-10} \specialrule{0em}{2pt}{1pt}
  	  &	\textbf{$\uparrow$ER} & \textbf{$\uparrow$EER$_1$} & \textbf{$\uparrow$EER$_2$} &	\textbf{$\uparrow$ER} & \textbf{$\uparrow$EER$_1$} & \textbf{$\uparrow$EER$_2$} &	\textbf{$\uparrow$ER} & \textbf{$\uparrow$EER$_1$} & \textbf{$\uparrow$EER$_2$}  \\
  \midrule[0.9pt]
  Bins ($b=1$) &1.000 & 0.254 & 0.292 & 1.000 & 0.226 & 0.287 & 1.000 & 0.425 & 0.373 \\
  Bins ($b=2$) &2.000 & 0.752 & 0.794 & 2.000 & 0.582 & 0.680 & 2.000 & 1.044 & 0.988 \\
  Bins ($b=3$) &3.000 & 1.137 & 1.266 & 3.000 & 0.999 & 1.089 & 3.000 & 1.683 & 1.605 \\
  Bins ($b=4$) &4.000 & 1.396 & 1.344 & 4.000 & 1.252 & 1.180 & 4.000 & 2.020 & 1.924 \\
  Bins ($b=5$) &5.000 & 1.245 & 1.280 & 5.000 & 1.230 & 1.180 & 5.000 & 2.500 & 2.135 \\
  \hline \specialrule{0em}{1pt}{1pt}
  Huffman ($k=1$) &1.000 & 0.218 & 0.219 & 1.000 & 0.219 & 0.231 & 1.000 & 0.430 & 0.387 \\
  Huffman ($k=2$) &1.824 & 0.593 & 0.600 & 1.824 & 0.546 & 0.635 & 1.841 & 0.924 & 0.893 \\
  Huffman ($k=3$) &2.509 & 1.024 & 1.202 & 2.518 & 0.927 & 1.083 & 2.595 & 1.638 & 1.549 \\
  Huffman ($k=4$) &3.145 & 1.809 & 1.950 & 3.224 & 1.496 & 1.821 & 3.266 & 2.387 & 2.404 \\
  Huffman ($k=5$) &3.705 & 2.427 & 2.627 & 3.872 & 2.249 & 2.602 & 3.932 & 3.133 & 3.200 \\
  \hline \specialrule{0em}{1pt}{1pt}
  Patient-Huffman ($\delta=1.0$) &1.125 & 0.927 & 0.949 & 0.809 & 0.713 & 0.740 & 0.988 & 0.933 & 0.886 \\
  Patient-Huffman ($\delta=1.5$) &1.711 & 1.182 & 1.083 & 1.460 & 0.952 & 0.925 & 1.668 & 1.369 & 1.400 \\
  Patient-Huffman ($\delta=2.0$) &2.129 & 1.184 & 1.220 & 1.905 & 1.050 & 1.044 & 2.201 & 1.541 & 1.490 \\
  \hline \specialrule{0em}{1pt}{1pt}
  Arithmetic ($h=100$) &4.224 & 3.371 & 3.527 & 4.412 & 3.269 & 3.459 & 4.308 & 3.908 & 3.843 \\
  Arithmetic ($h=200$) &4.651 & 4.051 & 4.125 & 4.908 & 3.981 & 4.324 & 4.805 & 4.449 & 4.219 \\
  Arithmetic ($h=300$) &4.903 & 4.207 & 4.290 & 5.127 & 4.532 & 4.450 & 4.942 & 4.630 & 4.606 \\
  \hline
  \hline \specialrule{0em}{1pt}{1pt}
  ADG &\textbf{5.147} & \textbf{4.648} & \textbf{4.699} & \textbf{5.650} & \textbf{5.164} & \textbf{5.435} & \textbf{5.411} & \textbf{5.373} & \textbf{5.384} \\
  \bottomrule[1.2pt]
  \end{tabular}
\end{table*}

\subsection{Implementation Details}
In experiments, we utilized LSTMs \cite{hochreiter1997long} for word-level generation. We stacked 2 LSTM layers and the model was implemented with Pytorch \cite{paszke2017automatic}. The dimension of word embedding was set to be 350. Hidden states in LSTM were set to be 512-dimensional vectors. In the training procedure, we applied SGD algorithm together with Adam \cite{kingma2014adam} to train the language model. Learning rate was set to be 0.001. The SGD update direction was computed using a batch of 32 training samples. They were both trained for 30 epochs on one GeForce GTX 1080 GPU. In the generation procedure, we adopted the model performing best on test sets. All generated sentences must be longer than 5 and shorter than 200.

\subsection{Baselines}
We rebuilt \citet{fang2017generating} (\textbf{Bins}), \citet{yang2018rnn} (\textbf{Huffman}), \citet{dai2019towards} (\textbf{Patient-Huffman}) and \citet{ziegler2019neural} (\textbf{Arithmetic}) as baselines. For fair comparison, we rebuilt all the baselines with the same language models. For Bins, we set $b$ to be 1, 2, 3, 4, 5 and the corresponding number of bins was 2, 4, 8, 16, 32. For Huffman, we built Huffman tree with the top 2, 4, 8, 16, 32 likely tokens. For Patient-Huffman, we measured the distortion by KL divergence and restricted the threshold $\delta$ to 1, 1.5, 2 with top 8 tokens. For Arithmetic, we truncated the conditional distribution at $h=100, 200, 300$. In each case, we generated 1,000 stegotext. We randomly chose same amount of covertext from the test sets for further evaluation.

\subsection{Metrics}
The metrics we utilized to evaluate the performance on embedding capacity and imperceptibility are listed as follows.

~\ \

\noindent \textbf{Embedding Rate} (\textbf{ER}): It is the average amount of information that one single token can carry, and is in unit of {\it bits per word} ({\it bpp}). Embedding rate is a metric to indicate the embedding capacity. Higher is better.

~\ \

\noindent \textbf{KL Divergence} between the implicit distribution $q$ and the modeled distribution $p_{LM}$ (\textbf{KLD$_1$}): It reflects the gap introduced by the embedding algorithm. Lower is better and the unit is {\it bit}. 

~\ \

\noindent \textbf{KL Divergence} between the statistical distributions of the sentence embedding of covertext and stegotext (\textbf{KLD$_2$}): It indirectly reflects the overall information-theoretic security. We mapped all stegotext and covertext to fixed length dense vectors $\bm{v_x}$ and $\bm{v_y}$ by third-party sentence vectorization tool \cite{le2014distributed}, and assumed that the resulting vectors of covertext and stegotext both obey isotropic Gaussian distribution. Then KLD$_2$ is computed by
\begin{equation}
  \begin{array}{ll}
    & D_{KL}(p(\bm{v_x}) || p(\bm{v_y}) ) \\
    &\approx \sum{(\log \frac{\bm{\sigma_{y}}}{\bm{\sigma_{x}}} + \frac{{\bm{\sigma_{x}}}^2 + (\bm{\mu_{x}} - \bm{\mu_{y}})^2}{2{\bm{\sigma_{y}}}^2} - \frac{1}{2})},
  \end{array}
\end{equation}
where $\bm{\mu}$ and $\bm{\sigma}$ are the mean and standard deviation of sentence vectors. We set the dimension of sentence vectors to be 100. Lower is better  and the unit is {\it bit}.

~\ \

\noindent \textbf{Detection Accuracy}: It reflects the anti-steganalysis ability of steganographic methods. Steganalysis is the technology used by Eve to detect hidden information in stegocarriers, which is the opposite direction of steganography. In our experiment, we employed linguistic steganalysis approaches based on Fasttext \cite{yang2019fast} (\textbf{Acc$_1$}) and TextCNN \cite{yang2020ts} (\textbf{Acc$_2$}). We took stegotext as positive samples and covertext as negative samples. We conducted 10-fold cross validation and reported the average accuracy. Closer to 50\% is better.

~\ \

\noindent \textbf{Effective Embedding Rate}: It is a new metric we proposed to evaluate the comprehensive performance of steganographic algorithms. It is defined to be calculated by
\begin{equation}
      \mbox{EER} = 2 \times (1-Acc) \times \mbox{ER},
\end{equation}
meaning that if the stegotext has a certain probability of being detected, the average amount of secret information actually transmitted should be discounted accordingly. For mathematical rigorousness and completeness, if $Acc < 0.5$, we assign $1 - Acc$ to $Acc$. In extreme cases where the stegocarriers are completely natural, the detection accuracy should be 50\% and EER is equal to ER. On the contrary, stegocarriers with 100\% detection accuracy cannot carry a single bit. We calculated this metric with the accuracy results obtained by the two aforementioned steganalysis method (\textbf{EER$_1$}, \textbf{EER$_2$}). Higher is better and the unit is {\it bpp}.

\begin{table*}[h]	
  \small
  \centering
  \caption{Examples of stegotext generated by ADG on the three corpora.}	
  \label{table result case}
 \begin{tabular}{m{24pt}|m{400pt}}
 \toprule[1.2pt]
 \multirow{5}{*}{\textbf{Movie}}& The supporting cast was also excellent. \\
 & But I guess you 've seen the many silent movies along with his other films. \\
 & And this movie was a precursor of val kilmer in the extreme. \\
 & It 's a unique wonderful movie that deserves all the recognition it deserved. \\
 & This is the worst movie I have ever seen. \\
 \hline \specialrule{0em}{1pt}{1pt}
 \multirow{5}{*}{\textbf{News}}& The FBI estimated its total wealth on Thursday. \\
 & Remember this is in part because of the actual policies of Donald Trump. \\
 & He said he did not care about any counterintelligence investigation. \\
 & Today however the process could not change even if he doesnt agree with Trumps rhetoric. \\
 & More than 100 000 people have been detained and another 30 000 civilians have been wounded early on Sunday. \\
 \hline \specialrule{0em}{1pt}{1pt}
 \multirow{5}{*}{\textbf{Tweet}}& Worst headache everrrr I dunno why but it was so scary. \\
 & I had a blast today in the MTV Movie Awards. \\
 & Ahhh some brothers do n't play sports! \\
 & Sadly you will be missing so much. \\
 & I do n't think the peach ice cream last night was good. \\
 \bottomrule[1.2pt]
  \end{tabular}
\end{table*}

\subsection{Results and Analysis}
The results of KLD$_1$ and KLD$_2$ are listed in Table \ref{table result kl}. KLD$_1$ measures the distortion between $q$ and $p_{LM}$, which is introduced by the embedding algorithm ADG. KLD$_2$ estimates the overall information-theoretic security that also considers the deviation of language models. In terms of KLD$_1$, we found that the proposed method ADG outperforms all baselines and it is very close to the optimal value 0 (stochastic sampling), which means generating stegotext by ADG is almost equivalent to normal generation with the language models. The results of KLD$_2$ are also advantageous, indicating that the generated stegotext is statistically consistent with the covertext. We noticed that some baselines can also perform well on KLD$_2$ (e.g. Patient-Huffman ($\delta=1.0$)). However, they have a crucial flaw in embedding capacity.

Table \ref{table result acc} demonstrates the results of anti-steganalysis, where we found the tendency coheres with that of KLD$_1$ and KLD$_2$. The proposed method ADG outperforms all baselines on the three corpora and it is very close to the optimal value 0.5, which further confirms its imperceptibility.
Besides, we also illustrated some examples of stegotext generated by ADG in Table \ref{table result case} for qualitative study. We found that the stegotext is fluent enough, with correct grammar and coherent semantics.

Finally, taking both embedding capacity and imperceptibility into account, we investigated effective embedding rate listed in Table \ref{table result eer}. It can be concluded that our method has excellent comprehensive performance, which outperforms all baselines. In general, the experimental results indicate that the proposed method ADG is able to resist both perceptual and statistical steganalysis of Eve, meanwhile ensure a remarkable embedding rate, which reveals its effectiveness.

\section{Conclusion}
\label{section conclusion}
Previous works of generative linguistic steganography inevitably introduce distortions to the distribution estimated by off-the-shelf language models.
In this paper, we attempted to achieve provably secure generative linguistic steganography during the procedure of stegotext generation. We proposed ADG, which embeds secret information by adaptive dynamic grouping. According to the mathematical proof and extensive experiments conducted on three public corpora, we found that the proposed method is provably secure and capable of generating fluent stegotext with high embedding capacity and high imperceptibility.
We hope our investigation of provably secure generative linguistic steganography can be leveraged as a building block for future research.

\section*{Acknowledgments}
This work is supported in part by the National Key Research and Development Program of China under Grant U1936216, Grant 61862002, Grant 6200197 and Grant U1705261. The authors thank anonymous reviewers for their insightful suggestions.


\bibliographystyle{acl_natbib}
\bibliography{anthology,acl2021}

\begin{thebibliography}{37}
\expandafter\ifx\csname natexlab\endcsname\relax\def\natexlab#1{#1}\fi

\bibitem[{Anderson and Petitcolas(1998)}]{anderson1998limits}
Ross~J Anderson and Fabien~AP Petitcolas. 1998.
\newblock On the limits of steganography.
\newblock \emph{IEEE Journal on selected areas in communications},
  16(4):474--481.

\bibitem[{Brown et~al.(2020)Brown, Mann, Ryder, Subbiah, Kaplan, Dhariwal,
  Neelakantan, Shyam, Sastry, Askell et~al.}]{brown2020language}
Tom~B Brown, Benjamin Mann, Nick Ryder, Melanie Subbiah, Jared Kaplan, Prafulla
  Dhariwal, Arvind Neelakantan, Pranav Shyam, Girish Sastry, Amanda Askell,
  et~al. 2020.
\newblock Language models are few-shot learners.
\newblock \emph{arXiv preprint arXiv:2005.14165}.

\bibitem[{Cachin(1998)}]{cachin1998information}
Christian Cachin. 1998.
\newblock An information-theoretic model for steganography.
\newblock In \emph{International Workshop on Information Hiding}, pages
  306--318. Springer.

\bibitem[{Chapman and Davida(1997)}]{chapman1997hiding}
Mark Chapman and George Davida. 1997.
\newblock Hiding the hidden: A software system for concealing ciphertext as
  innocuous text.
\newblock In \emph{International Conference on Information and Communications
  Security}, pages 335--345. Springer.

\bibitem[{Chapman and Davida(2002)}]{chapman2002plausible}
Mark Chapman and George Davida. 2002.
\newblock Plausible deniability using automated linguistic stegonagraphy.
\newblock In \emph{International Conference on Infrastructure Security}, pages
  276--287. Springer.

\bibitem[{Chapman et~al.(2001)Chapman, Davida, and
  Rennhard}]{chapman2001practical}
Mark Chapman, George~I Davida, and Marc Rennhard. 2001.
\newblock A practical and effective approach to large-scale automated
  linguistic steganography.
\newblock In \emph{International Conference on Information Security}, pages
  156--165. Springer.

\bibitem[{Dai and Cai(2019)}]{dai2019towards}
Falcon Dai and Zheng Cai. 2019.
\newblock Towards near-imperceptible steganographic text.
\newblock In \emph{Proceedings of the 57th Annual Meeting of the Association
  for Computational Linguistics}, pages 4303--4308.

\bibitem[{Dai et~al.(2010)Dai, Yu, Dai, and Deng}]{dai2010text}
Weihui Dai, Yue Yu, Yonghui Dai, and Bin Deng. 2010.
\newblock Text steganography system using markov chain source model and des
  algorithm.
\newblock \emph{JSW}, 5(7):785--792.

\bibitem[{Fang et~al.(2017)Fang, Jaggi, and Argyraki}]{fang2017generating}
Tina Fang, Martin Jaggi, and Katerina Argyraki. 2017.
\newblock Generating steganographic text with lstms.
\newblock In \emph{Proceedings of ACL 2017, Student Research Workshop}, pages
  100--106.

\bibitem[{Go et~al.(2009)Go, Bhayani, and Huang}]{go2009twitter}
Alec Go, Richa Bhayani, and Lei Huang. 2009.
\newblock Twitter sentiment classification using distant supervision.
\newblock \emph{CS224N project report, Stanford}, 1(12):2009.

\bibitem[{Hochreiter and Schmidhuber(1997)}]{hochreiter1997long}
Sepp Hochreiter and J{\"u}rgen Schmidhuber. 1997.
\newblock Long short-term memory.
\newblock \emph{Neural computation}, 9(8):1735--1780.

\bibitem[{Hussain et~al.(2018)Hussain, Wahab, Idris, Ho, and
  Jung}]{hussain2018image}
Mehdi Hussain, Ainuddin Wahid~Abdul Wahab, Yamani Idna~Bin Idris, Anthony~TS
  Ho, and Ki-Hyun Jung. 2018.
\newblock Image steganography in spatial domain: A survey.
\newblock \emph{Signal Processing: Image Communication}, 65:46--66.

\bibitem[{Jensen et~al.(1906)}]{jensen1906fonctions}
Johan Ludwig William~Valdemar Jensen et~al. 1906.
\newblock Sur les fonctions convexes et les in{\'e}galit{\'e}s entre les
  valeurs moyennes.
\newblock \emph{Acta mathematica}, 30:175--193.

\bibitem[{Kingma and Ba(2014)}]{kingma2014adam}
Diederik~P Kingma and Jimmy Ba. 2014.
\newblock Adam: A method for stochastic optimization.
\newblock \emph{arXiv preprint arXiv:1412.6980}.

\bibitem[{Krishnan et~al.(2017)Krishnan, Thandra, and
  Baba}]{krishnan2017overview}
R~Bala Krishnan, Prasanth~Kumar Thandra, and M~Sai Baba. 2017.
\newblock An overview of text steganography.
\newblock In \emph{2017 Fourth International Conference on Signal Processing,
  Communication and Networking (ICSCN)}, pages 1--6. IEEE.

\bibitem[{Le and Mikolov(2014)}]{le2014distributed}
Quoc Le and Tomas Mikolov. 2014.
\newblock Distributed representations of sentences and documents.
\newblock In \emph{International conference on machine learning}, pages
  1188--1196.

\bibitem[{Liu et~al.(2019)Liu, Liu, Wang, Zhao, and Liu}]{liu2019video}
Yunxia Liu, Shuyang Liu, Yonghao Wang, Hongguo Zhao, and Si~Liu. 2019.
\newblock Video steganography: A review.
\newblock \emph{Neurocomputing}, 335:238--250.

\bibitem[{Loper and Bird(2002)}]{loper2002nltk}
Edward Loper and Steven Bird. 2002.
\newblock Nltk: the natural language toolkit.
\newblock \emph{arXiv preprint cs/0205028}.

\bibitem[{Luo et~al.(2016)Luo, Huang, Li, and Chang}]{luo2016text}
Yubo Luo, Yongfeng Huang, Fufang Li, and Chinchen Chang. 2016.
\newblock Text steganography based on ci-poetry generation using markov chain
  model.
\newblock \emph{KSII Transactions on Internet and Information Systems (TIIS)},
  10(9):4568--4584.

\bibitem[{Maas et~al.(2011)Maas, Daly, Pham, Huang, Ng, and
  Potts}]{maas2011learning}
Andrew Maas, Raymond~E Daly, Peter~T Pham, Dan Huang, Andrew~Y Ng, and
  Christopher Potts. 2011.
\newblock Learning word vectors for sentiment analysis.
\newblock In \emph{Proceedings of the 49th annual meeting of the association
  for computational linguistics: Human language technologies}, pages 142--150.

\bibitem[{Mishra et~al.(2018)Mishra, Yadav, Trivedi, and
  Shrimali}]{mishra2018audio}
Shilpi Mishra, Virendra~Kumar Yadav, Munesh~Chandra Trivedi, and Tarun
  Shrimali. 2018.
\newblock Audio steganography techniques: A survey.
\newblock In \emph{Advances in Computer and Computational Sciences}, pages
  581--589. Springer.

\bibitem[{Moraldo(2014)}]{moraldo2014approach}
H~Hernan Moraldo. 2014.
\newblock An approach for text steganography based on markov chains.
\newblock \emph{arXiv preprint arXiv:1409.0915}.

\bibitem[{Paszke et~al.(2017)Paszke, Gross, Chintala, Chanan, Yang, DeVito,
  Lin, Desmaison, Antiga, and Lerer}]{paszke2017automatic}
Adam Paszke, Sam Gross, Soumith Chintala, Gregory Chanan, Edward Yang, Zachary
  DeVito, Zeming Lin, Alban Desmaison, Luca Antiga, and Adam Lerer. 2017.
\newblock Automatic differentiation in pytorch.

\bibitem[{Radford et~al.(2019)Radford, Wu, Child, Luan, Amodei, and
  Sutskever}]{radford2019language}
Alec Radford, Jeffrey Wu, Rewon Child, David Luan, Dario Amodei, and Ilya
  Sutskever. 2019.
\newblock Language models are unsupervised multitask learners.
\newblock \emph{OpenAI blog}, 1(8):9.

\bibitem[{Sharma et~al.(2016)Sharma, Gupta, Trivedi, and
  Yadav}]{sharma2016analysis}
Shivani Sharma, Avadhesh Gupta, Munesh~Chandra Trivedi, and Virendra~Kumar
  Yadav. 2016.
\newblock Analysis of different text steganography techniques: a survey.
\newblock In \emph{2016 Second International Conference on Computational
  Intelligence \& Communication Technology (CICT)}, pages 130--133. IEEE.

\bibitem[{Shirali-Shahreza(2008)}]{shirali2008text}
Mohammad Shirali-Shahreza. 2008.
\newblock Text steganography by changing words spelling.
\newblock In \emph{2008 10th international conference on advanced communication
  technology}, volume~3, pages 1912--1913. IEEE.

\bibitem[{Simmons(1984)}]{simmons1984prisoners}
Gustavus~J Simmons. 1984.
\newblock The prisoners’ problem and the subliminal channel.
\newblock In \emph{Advances in Cryptology}, pages 51--67. Springer.

\bibitem[{Wayner(1992)}]{wayner1992mimic}
Peter Wayner. 1992.
\newblock Mimic functions.
\newblock \emph{Cryptologia}, 16(3):193--214.

\bibitem[{Xiang et~al.(2018)Xiang, Li, Hao, Yang, and
  Shen}]{xiang2018reversible}
Lingyun Xiang, Yan Li, Wei Hao, Peng Yang, and Xiaobo Shen. 2018.
\newblock Reversible natural language watermarking using synonym substitution
  and arithmetic coding.
\newblock \emph{Comput. Mater. Continua}, 55(3):541--559.

\bibitem[{Xiang et~al.(2014)Xiang, Sun, Luo, and Xia}]{xiang2014linguistic}
Lingyun Xiang, Xingming Sun, Gang Luo, and Bin Xia. 2014.
\newblock Linguistic steganalysis using the features derived from synonym
  frequency.
\newblock \emph{Multimedia tools and applications}, 71(3):1893--1911.

\bibitem[{Yang et~al.(2018{\natexlab{a}})Yang, Guo, Chen, Huang, and
  Zhang}]{yang2018rnn}
Zhong-Liang Yang, Xiao-Qing Guo, Zi-Ming Chen, Yong-Feng Huang, and Yu-Jin
  Zhang. 2018{\natexlab{a}}.
\newblock Rnn-stega: Linguistic steganography based on recurrent neural
  networks.
\newblock \emph{IEEE Transactions on Information Forensics and Security},
  14(5):1280--1295.

\bibitem[{Yang et~al.(2020{\natexlab{a}})Yang, Zhang, Hu, Hu, and
  Huang}]{yang2020vae}
Zhong-Liang Yang, Si-Yu Zhang, Yu-Ting Hu, Zhi-Wen Hu, and Yong-Feng Huang.
  2020{\natexlab{a}}.
\newblock Vae-stega: Linguistic steganography based on variational
  auto-encoder.
\newblock \emph{IEEE Transactions on Information Forensics and Security},
  16:880--895.

\bibitem[{Yang et~al.(2019)Yang, Huang, and Zhang}]{yang2019fast}
Zhongliang Yang, Yongfeng Huang, and Yu-Jin Zhang. 2019.
\newblock A fast and efficient text steganalysis method.
\newblock \emph{IEEE Signal Processing Letters}, 26(4):627--631.

\bibitem[{Yang et~al.(2020{\natexlab{b}})Yang, Huang, and Zhang}]{yang2020ts}
Zhongliang Yang, Yongfeng Huang, and Yu-Jin Zhang. 2020{\natexlab{b}}.
\newblock Ts-csw: text steganalysis and hidden capacity estimation based on
  convolutional sliding windows.
\newblock \emph{Multimedia Tools and Applications}, 79(25):18293--18316.

\bibitem[{Yang et~al.(2018{\natexlab{b}})Yang, Jin, Huang, Zhang, and
  Li}]{yang2018automatically}
Zhongliang Yang, Shuyu Jin, Yongfeng Huang, Yujin Zhang, and Hui Li.
  2018{\natexlab{b}}.
\newblock Automatically generate steganographic text based on markov model and
  huffman coding.
\newblock \emph{arXiv preprint arXiv:1811.04720}.

\bibitem[{Zhou et~al.(2021)Zhou, Peng, Yang, Wen, Xue, and
  Zhong}]{zhou2021linguistic}
Xuejing Zhou, Wanli Peng, Boya Yang, Juan Wen, Yiming Xue, and Ping Zhong.
  2021.
\newblock Linguistic steganography based on adaptive probability distribution.
\newblock \emph{IEEE Transactions on Dependable and Secure Computing}.

\bibitem[{Ziegler et~al.(2019)Ziegler, Deng, and Rush}]{ziegler2019neural}
Zachary Ziegler, Yuntian Deng, and Alexander~M Rush. 2019.
\newblock Neural linguistic steganography.
\newblock In \emph{Proceedings of the 2019 Conference on Empirical Methods in
  Natural Language Processing and the 9th International Joint Conference on
  Natural Language Processing (EMNLP-IJCNLP)}, pages 1210--1215.

\end{thebibliography}
\end{document}